\documentclass{article} 
\usepackage{iclr2021_conference,times}

\usepackage{amsmath,amsfonts,bm}

\def\eqref#1{equation~\ref{#1}}

\def\1{\bm{1}}

\def\rx{{\textnormal{x}}}

\def\ve{{\bm{e}}}

\def\vg{{\bm{g}}}

\def\vk{{\bm{k}}}

\def\vq{{\bm{q}}}

\def\vx{{\bm{x}}}

\DeclareMathAlphabet{\mathsfit}{\encodingdefault}{\sfdefault}{m}{sl}
\SetMathAlphabet{\mathsfit}{bold}{\encodingdefault}{\sfdefault}{bx}{n}

\DeclareMathOperator*{\argmax}{arg\,max}
\DeclareMathOperator*{\argmin}{arg\,min}

\usepackage{hyperref}
\usepackage{url}
\usepackage[pdftex]{graphicx}
\usepackage{wrapfig}
\usepackage{multirow}
\usepackage{centernot}
\usepackage{caption}
\usepackage{subcaption}
\usepackage{outlines}
\usepackage{enumitem}
\usepackage{makecell}
\newcommand{\tabincell}[2]{\begin{tabular}{@{}#1@{}}#2\end{tabular}}

\newcommand{\ie}{\textit{i.e.}}
\newcommand{\eg}{\textit{e.g.}}

\title{CoDA: Contrast-enhanced and Diversity-promoting Data Augmentation for Natural Language Understanding}

\iclrfinalcopy

\author{Yanru Qu$^1$\thanks{Work was done during an internship at Microsoft Dynamics 365 AI.}, Dinghan Shen$^2$, Yelong Shen$^2$, Sandra Sajeev$^2$, Jiawei Han$^1$, Weizhu Chen$^2$ \\
$^1$University of Illinois, Urbana-Champaign, $^2$Microsoft Dynamics 365 AI\\
$^1$\texttt{\{yanruqu2,hanj\}@illinois.edu}, \\ $^2$\texttt{\{dishen,yeshe,ssajeev,wzchen\}@microsoft.com}
}

\begin{document}

\maketitle

\begin{abstract}



Data augmentation has been demonstrated as an effective strategy for improving model generalization and data efficiency. 
However, due to the discrete nature of natural language, designing label-preserving transformations for text data tends to be more challenging.
In this paper, we propose a novel data augmentation framework dubbed \emph{CoDA}, which synthesizes diverse and informative augmented examples by integrating multiple transformations organically.   
Moreover, a contrastive regularization objective is introduced to capture the \emph{global} relationship among all the data samples. 
A momentum encoder along with a memory bank is further leveraged to better estimate the contrastive loss. 
To  verify  the  effectiveness  of  the  proposed framework, we apply CoDA to Transformer-based models on a wide range of natural language understanding tasks.
On the GLUE benchmark, CoDA gives rise to an average improvement of $2.2\%$ while applied to the Roberta-large model. 
More importantly, it consistently exhibits stronger results relative to several competitive data augmentation and adversarial training baselines (including the low-resource settings).
Extensive experiments show that the proposed contrastive objective can be flexibly combined with various data augmentation approaches to further boost their performance, highlighting the wide applicability of the CoDA framework.  

\end{abstract}

\section{Introduction}\label{sec:intro}
\vspace{-3mm}

Data augmentation approaches have successfully improved large-scale neural-network-based models, 
\citep{laine2016temporal, xie2019unsupervised, berthelot2019mixmatch, sohn2020fixmatch, he2020momentum, khosla2020supervised, chen2020simple}, 
however, the majority of existing research is geared towards computer vision tasks.
The discrete nature of natural language makes it challenging to design effective label-preserving transformations for text sequences that can help improve model generalization \citep{hu2019learning, xie2019unsupervised}. 
On the other hand, fine-tuning powerful, over-parameterized language models\footnote{To name a few, BERT \citep{devlin2018bert}: 340M parameters, T5 \citep{raffel2019exploring}: 11B parameters, GPT-3 \citep{brown2020language}: 175B parameters.} proves to be difficult, especially when there is a limited amount of task-specific data available. It may result in representation collapse \citep{aghajanyan2020better} or require special finetuning techniques \citep{sun2019fine, hao2019visualizing}.
In this work, we aim to take a further step towards finding effective data augmentation strategies through systematic investigation. 

In essence, data augmentation can be regarded as constructing neighborhoods around a training instance that preserve the ground-truth label. 
With such a characterization, adversarial training \citep{zhu2019freelb, jiang2019smart, liu2020adversarial, cheng2020advaug} also performs label-preserving transformation in embedding space, and thus is considered as an alternative to data augmentation methods in this work.
From this perspective, the goal of developing effective data augmentation strategies can be summarized as answering three fundamental questions:

\emph{\romannumeral1}) What are some label-preserving transformations, that can be applied to text, to compose useful augmented samples?

\emph{\romannumeral2}) Are these transformations complementary in nature, and can we find some strategies to consolidate them for producing more diverse augmented examples?

\emph{\romannumeral3}) How can we incorporate the obtained augmented samples into the training process in an effective and principled manner?

Previous efforts in augmenting text data were mainly focused on answering the first question \citep{yu2018qanet, xie2019unsupervised, kumar2019submodular, wei2019eda, chen-etal-2020-mixtext, shen2020simple}. Regarding the second question, different label-preserving transformations have been proposed, but it remains unclear how to integrate them organically. In addition, it has been shown that the diversity of augmented samples plays a vital role in their effectiveness \citep{xie2019unsupervised, gontijo2020affinity}. 
In the case of image data, several strategies that combine different augmentation methods have been proposed, such as applying multiple transformations sequentially \citep{Cubuk2018AutoAugmentLA, cubuk2020randaugment, hendrycks2019augmix}, learning data augmentation policies \citep{Cubuk2018AutoAugmentLA}, randomly sampling operations for each data point \citep{cubuk2020randaugment}. However, these methods cannot be naively applied to text data, since the semantic meanings of a sentence are much more sensitive to local perturbations (relative to an image).

As for the third question, consistency training is typically employed to utilize the augmented samples \citep{laine2016temporal, hendrycks2019augmix, xie2019unsupervised, sohn2020fixmatch, miyato2018virtual}. This method encourages the model predictions to be invariant to certain label-preserving transformations. 
However, existing approaches only examine a pair of original and augmented samples in isolation, without considering other examples in the entire training set. 
As a result, the representation of an augmented sample may be closer to those of other training instances, rather than the one it is derived from. 
Based on this observation, we advocate that, in addition to consistency training, a training objective that can \emph{globally} capture the intrinsic relationship within the entire set of original and augmented training instances can help leverage augmented examples more effectively.

In this paper, we introduce a novel \textbf{Co}ntrast-enhanced and \textbf{D}iversity-promoting Data \textbf{A}ugmentation (CoDA) framework for natural language understanding. 
To improve the diversity of augmented samples, we extensively explore different combinations of isolated label-preserving transformations in an unified approach.  
We find that \emph{stacking} distinct label-preserving transformations produces particularly informative samples. 
Specifically, the most diverse and high-quality augmented samples are obtained by stacking an adversarial training module over the back-translation transformation. 
Besides the consistency-regularized loss for repelling the model to behave consistently within local neighborhoods, we propose a contrastive learning objective to capture the \emph{global} relationship among the data points in the representation space. 
We evaluate CoDA on the GLUE benchmark (with RoBERTa \citep{liu2019roberta} as the testbed), and CoDA consistently improves the generalization ability of resulting models and gives rise to significant gains relative to the standard fine-tuning procedure. Moreover, our method also outperforms various single data augmentation operations, combination schemes, and other strong baselines. Additional experiments in the low-resource settings and ablation studies further demonstrate the effectiveness of this framework.

\vspace{-2mm}
\section{Method}
\vspace{-2mm}
In this section, 
we focus our discussion on the natural language understanding (NLU) tasks, and particularly, under a text classification scenario. However, the proposed data augmentation framework can be readily extended to other NLP tasks as well.
\vspace{-2mm}
\subsection{Background: Data Augmentation and Adversarial Training}\label{sec:background}
\vspace{-2mm}
\paragraph{Data Augmentation} Let $\mathcal{D} = \{\vx_i, y_i\}_{i=1 \dots N}$ denote the training dataset, where the input example $\vx_i$ is a sequence of tokens, and $y_i$ is the corresponding label. 
To improve model's robustness and generalization ability, several data augmentation techniques (\eg, back-translation \citep{sennrich2015improving, edunov2018understanding, xie2019unsupervised}, mixup \citep{guo2019augmenting}, c-BERT \citep{wu2019conditional}) have been proposed. Concretely, label-preserving transformations are performed (on the original training sequences) to synthesize a collection of augmented samples, denoted by $\mathcal{D'} = \{\vx_i', y_i'\}_{i=1 \dots N}$ . 
Thus, a model can learn from both the training set $\mathcal{D}$ and the augmented set $\mathcal{D'}$, with $p_\theta(\cdot)$ the predicted output distribution of the model parameterized by $\theta$: 
\begin{align}
    \theta^{*} = \argmin_{\theta} \sum_{(\vx_i, y_i) \in \mathcal{D}} & \mathcal{L}\bigl(p_{\theta}(\vx_i), y_i\bigl) + \sum_{(\vx_i', y_i') \in \mathcal{D}'} \mathcal{L}\bigl(p_{\theta}(\vx_i'), y_i'\bigl)
\label{eq:aug}
\end{align}
Several recent research efforts were  focused on encouraging model predictions  to be invariant to stochastic or domain-specific data transformations \citep{xie2019unsupervised, laine2016temporal, tarvainen2017mean, sohn2020fixmatch, miyato2018virtual, jiang2019smart, hendrycks2019augmix}. 
Take back-translation as example: $\vx'_i = \text{BackTrans}(\vx_i)$, then $\vx'_i$ is a paraphrase of $\vx_i$. The model can be regularized to have consistent predictions for $(\vx_i, \vx'_i)$, by minimizing the distribution discrepancy $\mathcal{R}_{\text{CS}}(p_{\theta}(\vx_i), p_{\theta}({\vx}_i'))$, which typically adopts KL divergence (see Fig.~\ref{fig:back}).

\paragraph{Adversarial Training} In another line, adversarial training methods are applied to text data \citep{zhu2019freelb, jiang2019smart, cheng2020advaug, aghajanyan2020better} for improving model's robustness. Compared with data augmentation techniques, adversarial training requires no domain knowledge to generate additional training examples. Instead, it relies on the model itself to produce adversarial examples  which the model are most likely to make incorrect predictions. 
Similar to data augmentation, adversarial training also typically utilizes the cross-entropy and consistency-based objectives for training. As the two most popular adversarial-training-based algorithms, the adversarial loss \citep{goodfellow2014explaining} (Eqn. \ref{eq:at}) and virtual adversarial loss \citep{miyato2018virtual} (Eqn. \ref{eq:vat}) can be expressed as follows (see Fig.~\ref{fig:adv}):
\begin{align}
\mathcal{R}_{\text{AT}}(\vx_i, \tilde{\vx}_i, y_i) = \mathcal{L}\bigl(p_{\theta}(\tilde{\vx}_i),  y_i\bigl), s.t., \Vert \tilde{\vx}_i - \vx_i \Vert \leq \epsilon
\label{eq:at} ~,\\
\mathcal{R}_{\text{VAT}}(\vx_i, \tilde{\vx}_i) =  \mathcal{R}_{\text{CS}}\bigl(p_{\theta}(\tilde{\vx}_i),  p_{\theta}({\vx_i}) \bigl), s.t., \Vert \tilde{\vx}_i - \vx_i \Vert \leq \epsilon \label{eq:vat} ~.
\end{align}
Generally, there is no closed-form to obtain the exact adversarial example $\hat{\vx}_i$ in either Eqn. \ref{eq:at} or \ref{eq:vat}. However, it usually can be approximated by a low-order approximation of the objective function with respect to $\vx_i$. For example, the adversarial example in Eqn. \ref{eq:at} can be approximated by:
\begin{align}
    \hat{\vx}_i \approx \vx_i + \epsilon \frac{\vg}{\Vert \vg \Vert_2}, \text{where  } \vg = \nabla_{\vx_i} \mathcal{L}\bigl( p_{\theta}({\vx_i}), y_i \bigl) ~.
\end{align}

\vspace{-2mm}
\subsection{Diversity-promoting Consistency Training}\label{sec:stack}
\vspace{-2mm}

\begin{figure}[t]
\begin{center}
\vspace{-2mm}
\begin{subfigure}[t]{0.5\textwidth}
\centering
\includegraphics[scale=0.4]{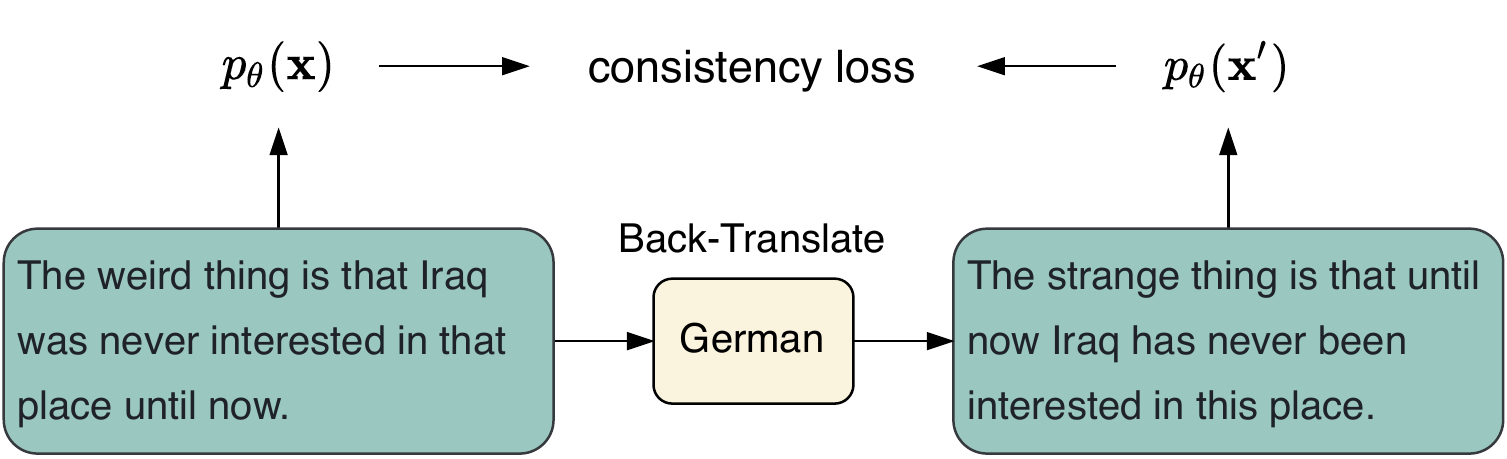}
\caption{Back-translation} \label{fig:back}
\end{subfigure}
\begin{subfigure}[t]{0.45\textwidth}
\centering
\includegraphics[scale=0.4]{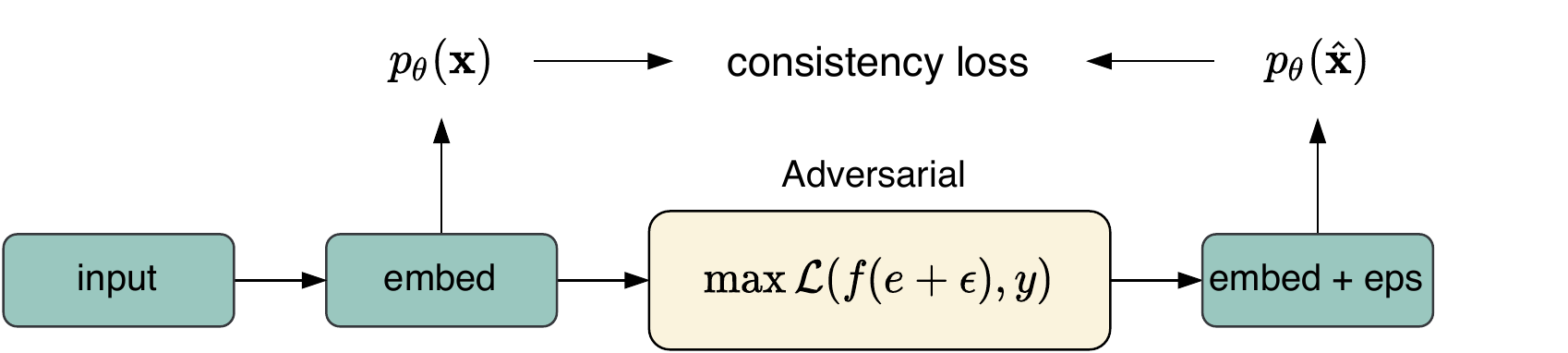}
\caption{Adversarial training} \label{fig:adv}
\end{subfigure}
\begin{subfigure}[t]{0.8\textwidth}
\centering
\includegraphics[scale=0.4]{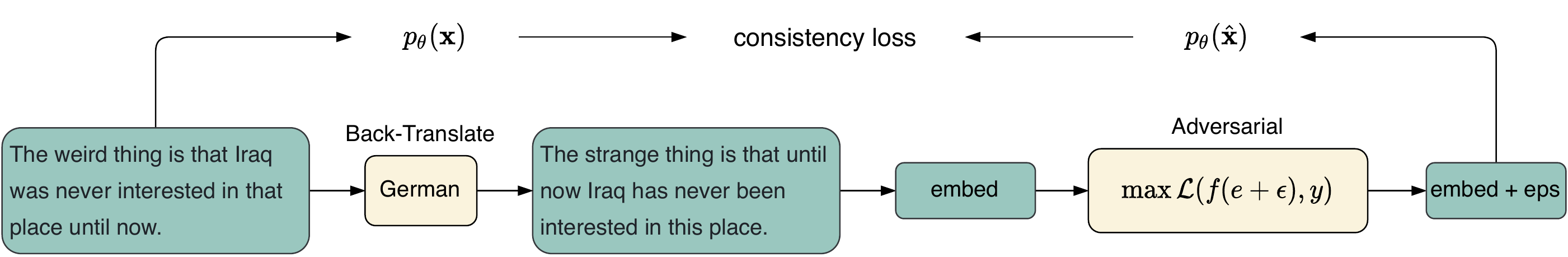}
\caption{Stacking of back-translation and adversarial training}
\label{fig:stack}
\end{subfigure}
\end{center}
\vspace{-2mm}
\caption{Illustration of data augmentation combined with adversarial training.}
\vspace{-2mm}
\end{figure}

As discussed in the previous section, data augmentation and adversarial training share the same intuition of producing neighbors around the original training instances. Moreover, both approaches share very similar training objectives. Therefore, it is natural to ask the following question: are different data augmentation methods and adversarial training equal in nature? Otherwise, are they complementary to each other, and thus can be consolidated together to further improve the model's generalization ability? Notably, it has been shown, in the CV domain, that combining different data augmentation operations could lead to more diverse augmented examples \citep{Cubuk2018AutoAugmentLA, cubuk2020randaugment, hendrycks2019augmix}. However, this is especially challenging for natural language, given that the semantics of a sentence can be entirely altered by slight perturbations.



\begin{figure}[t]
    \centering
     \begin{subfigure}[b]{0.35\textwidth}
     \centering
     \includegraphics[scale=0.4]{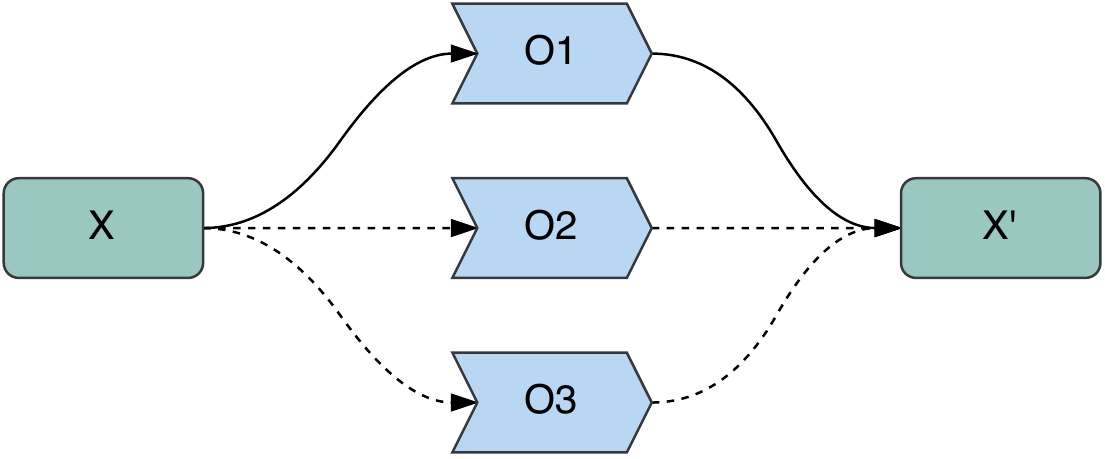}
     \caption{Random combination}
     \end{subfigure}
     \begin{subfigure}[b]{0.3\textwidth}
     \centering
     \includegraphics[scale=0.4]{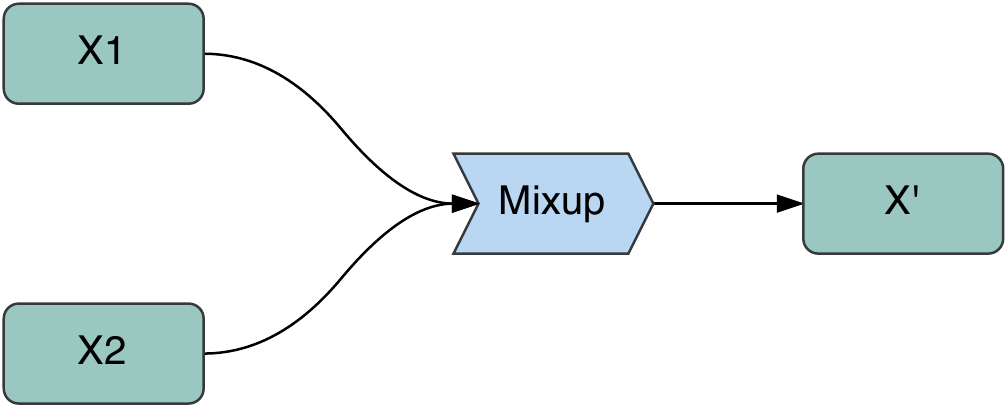}
     \caption{Mixup interpolation}
    \end{subfigure}
    \begin{subfigure}[b]{0.3\textwidth}
     \centering
     \includegraphics[scale=0.4]{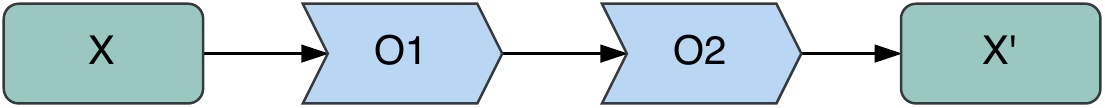}
     \caption{Sequential stacking}
     \label{fig:seq_stack}
    \end{subfigure}
    \caption{Illustration of different strategies to combine various label-preserving transformations.}
    \vspace{-3mm}
    \label{fig:comb}
\end{figure}

To answer the above question, we propose several distinct strategies to combine different data transformations, with the hope to produce more diverse and informative augmented examples. 
Specifically, we consider $5$ different types of label-preserving transformations: \emph{back-translation} \citep{sennrich2015improving, edunov2018understanding, xie2019unsupervised}, \emph{c-BERT} word replacement \citep{wu2019conditional}, \emph{mixup} \citep{guo2019augmenting, chen-etal-2020-mixtext}, \emph{cutoff} \citep{shen2020simple}, and \emph{adversarial training} \citep{zhu2019freelb, jiang2019smart}.
The $3$ combination strategies are schematically illustrated in Figure~\ref{fig:comb}. For random combination, a particular label-preserving transformation is randomly selected, among all the augmentation operations available, for each mini-batch. As to the mixup interpolation, given two samples $\vx_i$ and $\vx_j$ drawn in a mini-batch, linear interpolation is performed between their input embedding matrices $\ve_i$ and $\ve_j$ \citep{zhang2017mixup}: $\ve'_i = a \ve_i + (1 - a) \ve_j$, where $a$ is the interpolation parameter, usually drawn from a \emph{Beta} distribution.

Moreover, we consider stacking different label-preserving transformations in a sequential manner (see Figure~\ref{fig:seq_stack}). It is worth noting that due to the discrete nature of text data, some stacking orders are infeasible. For example, it is not reasonable to provide an adversarially-perturbed embedding sequence to the back-translation module.
Without loss of generality, we choose the combination where adversarial training is stacked over back-translation to demonstrate the sequential stacking operation (see Fig.~\ref{fig:stack}). Formally, given a training example $(\vx_i, y_i)$, 
the consistency training objective for such a stacking operation can be written as: 
\begin{align}
{\vx}_i' = \text{BackTrans}(\vx_i), \text{ }\hat{\vx}_i & \approx \text{argmax}_{\tilde{\vx}_i} \mathcal{R}_{\text{AT}}(\vx_i', \tilde{\vx}_i, y_i) \label{eq:stack_adv} ~, \\
\mathcal{L}_{\text{consistency}}(\vx_i, \hat\vx_i, y_i) = \mathcal{L}\bigl(p_\theta(\vx_i), y_i\bigl) + & \alpha 
\mathcal{L}(p_\theta(\hat{\vx}_i), y_i)
+ \beta\mathcal{R}_{\text{CS}}(p_{\theta}(\vx_i), p_{\theta}(\hat{\vx}_i)) \label{eq:stack_loss} ~,
\end{align}
where the first term correspond to the cross-entropy loss, the second term is the adversarial loss, $\mathcal{R}_{\text{CS}}$ denotes the consistency loss term between $(\vx_i, \hat{\vx}_i)$. Note that $\hat{\vx}_i$ is obtained through two different label-preserving transformations applied to $\vx$, and thus deviates farther from $\vx$ and should be more \emph{diverse} than $\vx_i'$. Inspired by \citep{bachman2014learning, zheng2016improving, kannan2018adversarial, hendrycks2019augmix}, we employ the Jensen-Shannon divergence for $\mathcal{R}_{\text{CS}}$, since it is upper bounded and tends to be more stable and consistent relative to the KL divergence:
\begin{align}
\mathcal{R}_{\text{CS}}(p_{\theta}(\vx_i), p_{\theta}(\hat{\vx}_i)) = \frac{1}{2}\bigl(\text{KL}(p_{\theta}(\vx_i) \Vert M) + \text{KL}(p_{\theta}(\hat{\vx}_i)) \Vert M)\bigl) ~,
\end{align}
where $M = (p_{\theta}(\vx_i) + p_{\theta}(\hat{\vx}_i)) / 2$. Later we simply use $\vx'_i$ to represent the transformed example.





\subsection{Contrastive Regularization}\label{sec:contrastive}
\begin{wrapfigure}{r}{0.45\textwidth}
\vspace{-15mm}
\centering
\includegraphics[scale=0.5]{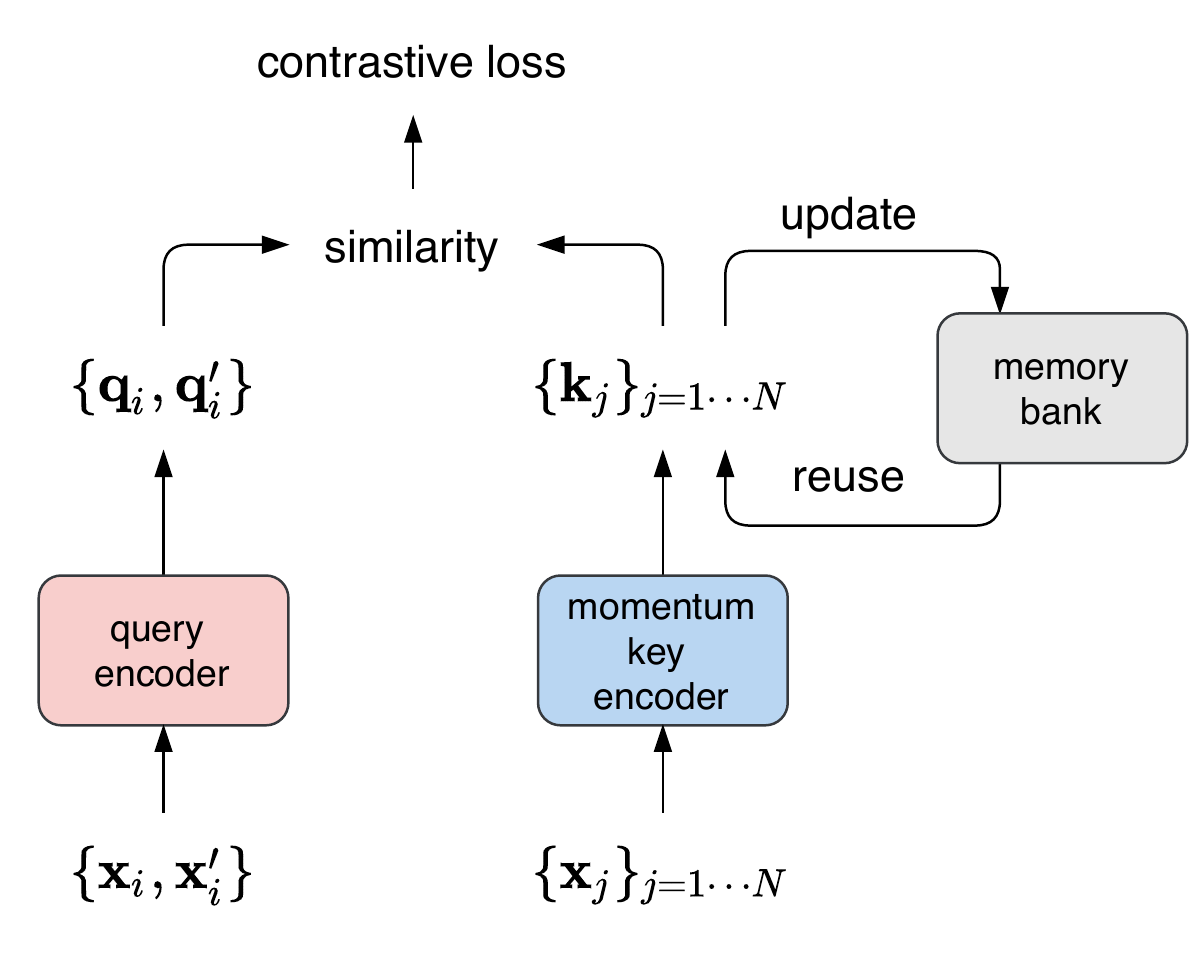}
\caption{Illustration of the contrastive learning module.}
\label{fig:teaser}
\end{wrapfigure}

Consistency loss only provides local regularization, \emph{i.e.}, $\vx_i$ and $\vx'_i$ should have close predictions. However, the relative positions between $\vx'_i$ and other training instances $\vx_j$ ($j\neq i$) have not been examined. 
In this regard, we propose to leverage a contrastive learning objective to better utilize the augmented examples.
Specifically, we assume that the model should encourage an augmented sample $\vx'_i$ to be closer, in the representation space, to its original sample $\vx_i$, relative to other data points $\vx_{j}$ ($j\neq i$) in the training set.
This is a reasonable assumption since intuitively, the model should be robust enough to successfully determine from which original data an augmented sample is produced.

The contrastive learning module is illustrated in Fig.~\ref{fig:teaser}. 
As demonstrated by prior efforts on contrastive learning, adopting a large batch size is especially vital for its effectiveness \citep{chen2020simple, khosla2020supervised}. 
Therefore, we introduce a memory bank that stores the history embeddings, thus enabling much larger number of negative samples. 
Moreover, to avoid the encoder from changing too rapidly (which may result in inconsistency embeddings), a momentum encoder module is incorporated into our algorithm. 
Concretely, let $f_{\theta}(.)$ and ${f}_{\bar{\theta}}(.)$ denote the transformation parameterized by the query encoder and key encoder, respectively. Note that $\theta$ and $\bar{\theta}$ represent their parameters. 
The momentum model parameters $\bar{\theta}$ are not learned by gradients. Instead, they are updated through the momentum rule: $\bar{\theta} \leftarrow \gamma \bar{\theta} + (1- \gamma) \theta$ at each training step. We omit the details here and refer the interested readers to the work by \citep{he2020momentum} for further explanation. 
Given a sample $\vx_i$ and its augmented example $\vx_i'$, the query and key can be obtained as follows:
\begin{align}
\vq_i = f_{\theta}(\vx_i), \quad \vq_i' = f_{\theta}(\vx_i'), \quad \vk_i = f_{\bar{\theta}}(\vx_i)~. 
\end{align}
Thus, the contrastive training objective can be written as: 
\begin{align}
\mathcal{R}_{\text{contrast}}(\vx_i,\vx_i',\mathcal{M}) & = 
\mathcal{R}_{\text{CT}}(\vq_i, \vk_i, \mathcal{M}) + \mathcal{R}_{\text{CT}}(\vq_i', \vk_i, \mathcal{M})
\label{eq:full_ct_1oss}, \\
\mathcal{R}_{\text{CT}}(\vq_i, \vk_i, \mathcal{M}) & = -\text{log}\frac{\text{exp}({\text{sim}(\vq_i, \vk_i) / \tau})}{\sum_{\vk_j \in \mathcal{M} \bigcup \{\vk_i\} }\text{exp}({\text{sim}(\vq_i, \vk_j) / \tau}) },  \label{eq:r_ct}
\end{align}
where $\tau$ is the temperature, and $\mathcal{M}$ is the memory bank in which the history keys are stored. 
Cosine similarity is chosen for $\text{sim}(\cdot)$.
Note that $\mathcal{R}_{\text{CT}}(\vq_i', \vk_i, \mathcal{M})$ is similarly defined as $\mathcal{R}_{\text{CT}}(\vq_i, \vk_i, \mathcal{M})$ (with $\vq_i$ replaced by $\vq_i'$ in Eqn.~\ref{eq:r_ct}).
In Eqn.~\ref{eq:full_ct_1oss}, the first term corresponds to the contrastive loss calculated on the original examples (self-contrastive loss), while the second term is computed on the augmented sample (augment-contrastive loss).  
Under such a framework, the pair of original and augmented samples are encouraged to stay closer in the learned embedding space, relative to all other training instances. As a result, the model is regularized \emph{globally} through considering the embeddings of all the training examples available.

By integrating both the consistency training objective and the contrastive regularization, the overall training objective for the CoDA framework can be expressed as:
\begin{align}
    \theta^{*} = \text{argmin}_{\theta}\sum_{(\vx_i, y_i) \in \mathcal{D}}\mathcal{L}_{\text{consistency}}(\vx_i, \vx_i', y_i)+ \lambda \mathcal{R}_{\text{contrast}}\bigl(\vx_i, \vx_i', \mathcal{M} \bigl)
    \label{eq:objective}~.
\end{align}
where $\lambda$ is a hyperparameter to be chosen. It is worth noting that the final objective has taken both the \emph{local} (consistency loss) and \emph{global} (contrastive loss) information introduced by the augmented examples into consideration.

\vspace{-2mm}
\section{Experiments}
\vspace{-2mm}

To verify the effectiveness of \emph{CoDA}, We evaluate it on the widely-adopted GLUE benchmark \citep{wang2018glue}, which consists of multiple natural language understanding (NLU) tasks. The details  of these datasets can be found in Appendix ~\ref{ap:data}. 
RoBERTa \citep{liu2019roberta} is employed as the testbed for our experiments. However, the proposed approach can be flexibly integrated with other models as well. We provide more implementation details in Appendix~\ref{ap:implement}. Our
code will be released to encourage future research.

In this section, we first present our exploration of several different strategies to consolidate various data transformations (Sec~\ref{sec:da_design}). 
Next, we conduct extensive experiments to carefully select the contrastive objective for NLU problems in Sec~\ref{sec:contr_design}.
Based upon these settings, we further evaluate CoDA on the GLUE benchmark and compare it with a set of competitive baselines in Sec~\ref{sec:glue}.
Additional experiments in the low-resource settings and qualitative analysis (Sec~\ref{sec:analysis}) are further conducted to gain a deep understanding of the proposed framework.
\vspace{-2mm}
\subsection{Combining Label-preserving Transformations}\label{sec:da_design}
\vspace{-2mm}

We start by implementing and comparing several data augmentation baselines. As described in the previous section, we explore $5$ different approaches: \emph{back-translation}, \emph{c-BERT} word replacement, \emph{Mixup}, \emph{Cutoff} and \emph{adversarial training}. More details can be found in Appendix~\ref{ap:da}. The standard cross-entropy loss, along with the consistency regularization term (Eq.~\ref{eq:stack_loss}) is utilized for all methods to ensure a fair comparison. We employ the MNLI dataset and RoBERTa-base model for the comparison experiments with the results shown in Table~\ref{tab:da}.

All these methods have achieved improvements over the RoBERTa-base model, demonstrating the effectiveness of leveraging label-preserving transformations for NLU. Moreover, back-translation, cutoff and adversarial training exhibit stronger empirical results relative to mixup and c-BERT.

To improve the diversity of augmented examples, we explore several strategies to combine multiple transformations: \emph{\romannumeral1}) random combination, \emph{\romannumeral2}) mixup interpolation, and \emph{\romannumeral3}) sequential stacking, as shown in Fig.~\ref{fig:comb}. In Table~\ref{tab:da}, the score of naive random combination lies between single transformations. 
This may be attributed to the fact that different label-preserving transformations regularize the model in distinct ways, and thus the model may not be able to leverage different regularization terms simultaneously.

\begin{wraptable}{r}{0.5\linewidth}
    \vspace{-2mm}
    \small
    \centering
    \begin{tabular}{l|c|c} \hline
        \textbf{Method} & \tabincell{c}{MNLI-m\\(Acc)} & MMD \\ \hline \hline
        RoBERTa-base & 87.6 & - \\ \hline \hline
        \multicolumn{3}{c}{\textbf{\emph{Single Transformation}}} \\ \hline
        + back-translation & 88.5 &  0.63 \\ 
        + c-BERT & 88.0 & 0.01 \\
        + cutoff & 88.4 & 0.02 \\
        + mixup (ori, ori) & 88.2 & 0.06 \\
        + adversarial & 88.5 & 0.65 \\ \hline \hline
        \multicolumn{3}{c}{\textbf{\emph{Multiple Transformations}}} \\ \hline
        + random (back, cut, adv) & 88.4 & - \\
        + mix (ori, back) & 88.4 & 0.11 \\
        + mix (back, adv) & 88.6 & 0.81 \\
        + stack (back, cut) & 88.5 & 0.62 \\
        + stack (back, adv) & \textbf{88.8} & 1.14 \\
        + stack (back, cut, adv) & 88.5 & 1.14 \\
        + stack (back, adv, cut) & 88.4 & 1.14 \\ \hline
    \end{tabular}
    \caption{Comparison of different transformations on the MNLI-m development set. \emph{Abbr:} original training instances (ori), back-translation (back), cutoff (cut), mixup (mix), adversarial (adv).}
    \label{tab:da}
    \vspace{-6mm}
\end{wraptable}

Besides, among all the other combination strategies, we observe that gains can be obtained by integrating back-translation and adversarial training together. Concretely, mixing back-translation and adversarial training samples (in the input embedding space) slightly improve the accuracy from $88.5$ to $88.6$. More importantly, the result is further improved to $88.8$ with these two transformations stacking together\footnote{In practice, we can use Machine Translation (MT) models trained on large parallel corpus (\eg, English-French, English-German) to back-translate the input sentence. Since back-translation requires decoding, it can be performed offline. If the input contains multiple sentences, we split it into sentences, perform back-translation, and ensemble those paraphrases back.} (see Sec~\ref{sec:stack}). With significance test, we find \emph{stack (back, adv)} performs consistently better than other combinations. This observation indicates that the \emph{stacking} operation, especially in the case of back-translation and adversarial training, can produce more diverse augment examples. 

Intuitively, the augmented sample, with two sequential transformations, deviates more from the corresponding training data, and thus tends to be more effective at improving the model's generalization ability. 
To verify this hypothesis, we further calculate the MMD \citep{gretton2012kernel} between augmented samples and the original training instances. It can be observed that \emph{stack (back, adv)}, \emph{stack (back, cut, adv)} and \emph{stack (back, adv, cut)} have all produced examples the farthest from the original training instances (see Table~\ref{tab:da}). However, we conjecture that the latter two may have altered the semantic meanings too much, thus leading to inferior results. 
In this regard, \emph{stack (back, adv)} is employed as the data transformation module for all the experiments below.


\vspace{-3mm}
\subsection{Contrastive Regularization Design}\label{sec:contr_design}
\begin{figure}[ht]
     \centering
     \vspace{-5mm}
     \begin{subfigure}[b]{0.45\textwidth}
         \centering
         \includegraphics[scale=0.35]{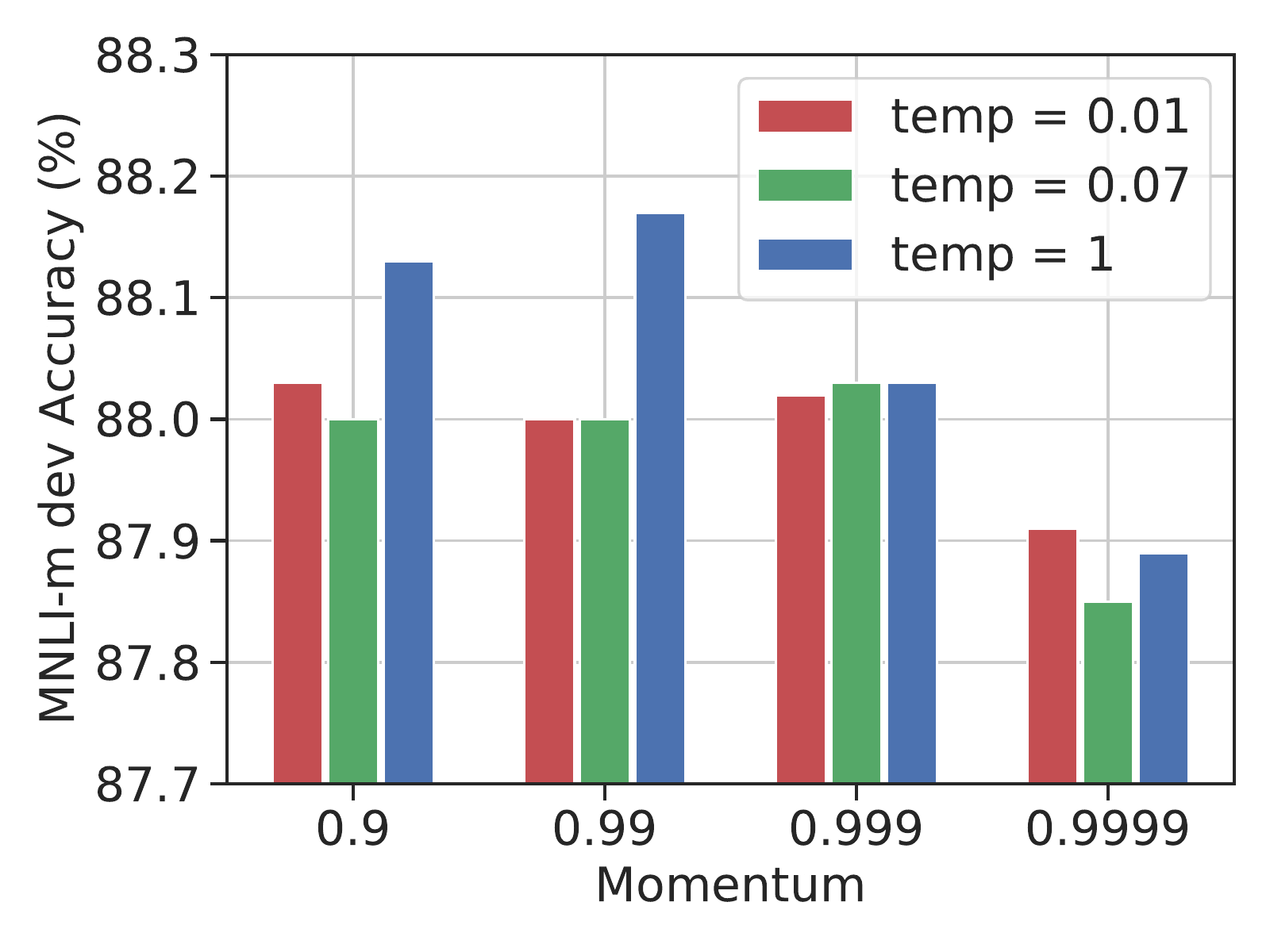}
         \vspace{-2mm}
         \caption{Momentum encoder}
         \label{fig:temp_mom}
     \end{subfigure}
     \begin{subfigure}[b]{0.45\textwidth}
         \centering
         \includegraphics[scale=0.35]{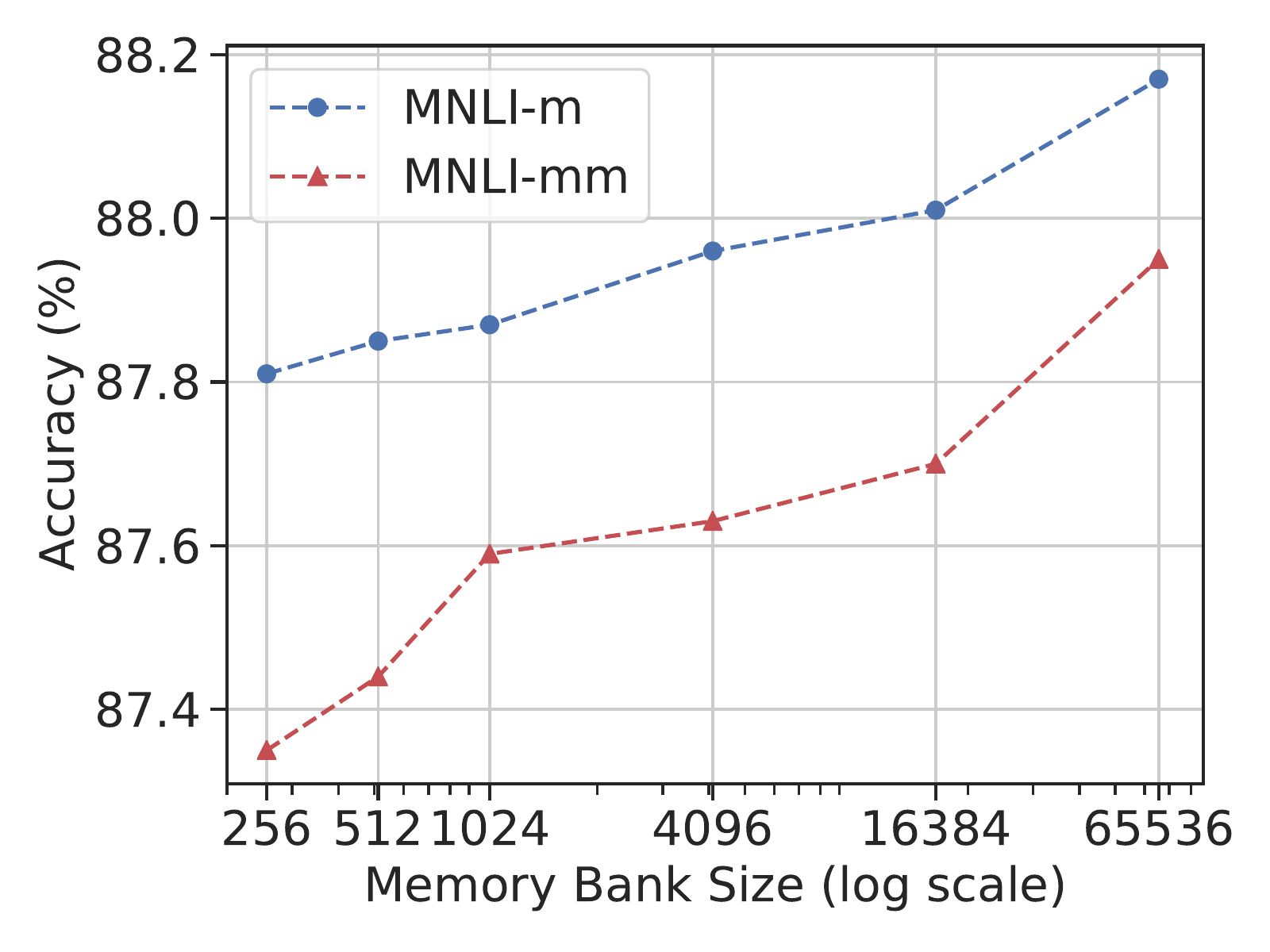}
         \vspace{-2mm}
         \caption{Memory bank}
         \label{fig:mem}
     \end{subfigure}
     \vspace{-2mm}
    \caption{Hyperparameter exploration for the contrastive loss, evaluated on the MNLI-m development set. \emph{Note:} All models use the RoBERTa-base model as the encoder.}
    \vspace{-2mm}
    \label{}
\end{figure}

\vspace{-3mm}
In this section, we aim to incorporate the \emph{global} information among the entire set of original and augmented samples \emph{via} a contrastive regularization. 
First, we explore a few hyperparameters for the proposed contrastive objective. Since both the memory bank and the momentum encoder are vital components, we study the impacts of different hyperparameter values on both the temperature and the momentum. As shown in Fig.~\ref{fig:temp_mom}, a temperature of $1.0$ combined with the momentum of $0.99$ can achieve the best empirical result. We then examine the size effect of the memory bank, and observe a larger memory bank size leads to a better capture of the global information and results in higher performance boost\footnote{We set the default memory bank size as $65536$. As to smaller datasets, we choose the size no larger than the number of training data (\eg, MRPC has $3.7$k examples, and we set the memory size as $2048$).} (see Fig.~\ref{fig:mem}).

After carefully choosing the best setting based on the above experiments, we apply the contrastive learning objective to several GLUE datasets. We also implement several prior works on contrastive learning to compare, including the MoCo loss \citep{he2020momentum} and the supervised contrastive (SupCon) loss \citep{khosla2020supervised}, all implemented with memory banks. Note that we remove the consistency regularization for this experiment to better examine the effect of the contrastive regularization term (\ie, $\alpha = \beta = 0$, $\lambda \neq 0$). As presented in Table~\ref{tab:contr}, our contrastive objective consistently exhibits the largest performance improvement. This observation demonstrates for NLU, our data transformation module can be effectively equipped with the contrastive regularization.  


\vspace{-2mm}
\begin{table}[h] 
    \small
    \centering
    \setlength{\tabcolsep}{4.5pt}
    \def\arraystretch{1.0}
    \begin{tabular}{l|c|c|c|c|c} \hline
        \multirow{2}{*}{\textbf{Method}} & MNLI-m & QNLI & SST-2 & RTE & MRPC \\
        & (Acc) & (Acc) & (Acc) & (Acc) & (Acc) \\ \hline \hline
        RoBERTa-base & 87.6	& 92.8	& 94.8	& 78.7	& 90.2 \\ \hline
        + MoCo \citep{he2020momentum} & \textbf{88.2}	& 93.3	& 95.1	& 80.8	& 90.9 \\ 
        + SupCon \citep{khosla2020supervised} & 88.1	& 93.2	& 95.2	& 80.5	& 90.2 \\ 
        + Contrastive (ours) & 88.1 & \textbf{93.6} & \textbf{95.3} & \textbf{82.0} & \textbf{91.7} \\ \hline
    \end{tabular}
    \vspace{-2mm}
    \caption{Comparison among different contrastive objectives on the GLUE development set.}
    \vspace{-4mm}
    \label{tab:contr}
\end{table}

\vspace{-1mm}
\subsection{GLUE Benchmark Evaluation}\label{sec:glue}
\vspace{-2mm}
With both components within the CoDA algorithm being specifically tailored to the natural language understanding applications, we apply it to the RoBERTa-large model \citep{liu2019roberta}. Comparisons are made with several competitive data-augmentation-based and adversarial-training-based approaches on the GLUE benchmark. Specifically, we consider back-translation, cutoff \citep{shen2020simple}, FreeLB \citep{zhu2019freelb}, SMART \citep{jiang2019smart}, and R3F \citep{aghajanyan2020better} as the baselines, where the last three all belong to adversarial training. The results are presented in Table~\ref{tab:glue}. It is worth noting that back-translation is based on our implementation, where both the cross-entropy and consistency regularization terms are utilized.

\begin{table}[h]
    \small
    \centering
    \setlength{\tabcolsep}{4.5pt}
    \def\arraystretch{1.0}
    \vspace{-1mm}
    \begin{tabular}{l|c|c|c|c|c|c|c|c|c}
        \hline \textbf{Method} & \tabincell{c}{MNLI-m/\\mm (Acc)} & \tabincell{c}{QQP\\(Acc/F1)} & \tabincell{c}{QNLI\\(Acc)} & \tabincell{c}{SST-2\\(Acc)} & \tabincell{c}{MRPC\\(Acc/F1)} & \tabincell{c}{CoLA\\(Mcc)} & \tabincell{c}{RTE\\(Acc)} & \tabincell{c}{STS-B\\(P/S)} & Avg \\ \hline \hline
        RoBERTa-large & 90.2/-	& 92.2/-	& 94.7	& 96.4	& -/90.9	& 68	& 86.6	& 92.4/- & 88.9\\
        \hline
        Back-Trans & 91.1/90.4	& 92/-	& 95.3	& 97.1	& 90.9/93.5	& 69.4	& 91.7	& 92.8/92.6 & 90.4 \\
        Cutoff &  91.1/- & 92.4/-  & 95.3 & 96.9 & 91.4/93.8 & 71.5 & 91.0  & 92.8/- & 90.6 \\
        FreeLB &        90.6/-	& \textbf{92.6}/-	& 95	& 96.7	& 91.4/-	& 71.1	& 88.1	& 92.7/- & - \\
        SMART & 91.1/\textbf{91.3}	& 92.4/89.8	& \textbf{95.6}	& 96.9	& 89.2/92.1	& 70.6	& 92	& 92.8/92.6 & 90.4 \\
        R3F & 91.1/\textbf{91.3} & 92.4/\textbf{89.9} & 95.3 & 97.0 & 91.6/- & 71.2 & 88.5 & - & - \\ 
        \hline
        CoDA & \textbf{91.3}/90.8 & 92.5/\textbf{89.9} & 95.3 & \textbf{97.4} & \textbf{91.9/94} & \textbf{72.6} & \textbf{92.4} & \textbf{93/92.7} & \textbf{91.1} \\ \hline
    \end{tabular}
    \vspace{-1mm}
    \caption{Main results of single models on the GLUE development set. \emph{Note:} The best result on each task is in \textbf{bold} and ``-'' denotes the missing results. The average score is calculated based on the same setting as RoBERTa.}
    \vspace{-3mm}
    \label{tab:glue}
\end{table}

We find that \emph{CoDA} brings significant gains to the RoBERTa-large model, with the averaged score on the GLUE dev set improved from $88.9$ to $91.1$. More importantly, CoDA consistently outperforms these strong baselines (indicated by a higher averaged score), demonstrating that our algorithm can produce informative and high-quality augmented samples and leverage them effectively as well. Concretely, on datasets with relatively larger numbers of training instances ($>$ 100K), \emph{i.e.}, MNLI, QQP and QNLI, different approaches show similar gains over the RoBERTa-large model. However, on smaller tasks (SST-2, MRPC, CoLA, RTE, and STS-B), CoDA beats other data augmentation or adversarial-based methods by a wide margin. We attribute this observation to the fact that the synthetically produced examples are more helpful when the tasks-specific data is limited. Thus, when smaller datasets are employed for fine-tuning large-scale language models, the superiority of the proposed approach is manifested to a larger extent.

\begin{figure}[h]
 \centering
 \vspace{-4mm}
 \begin{subfigure}[]{0.45\textwidth}
     \centering
     \includegraphics[scale=0.35]{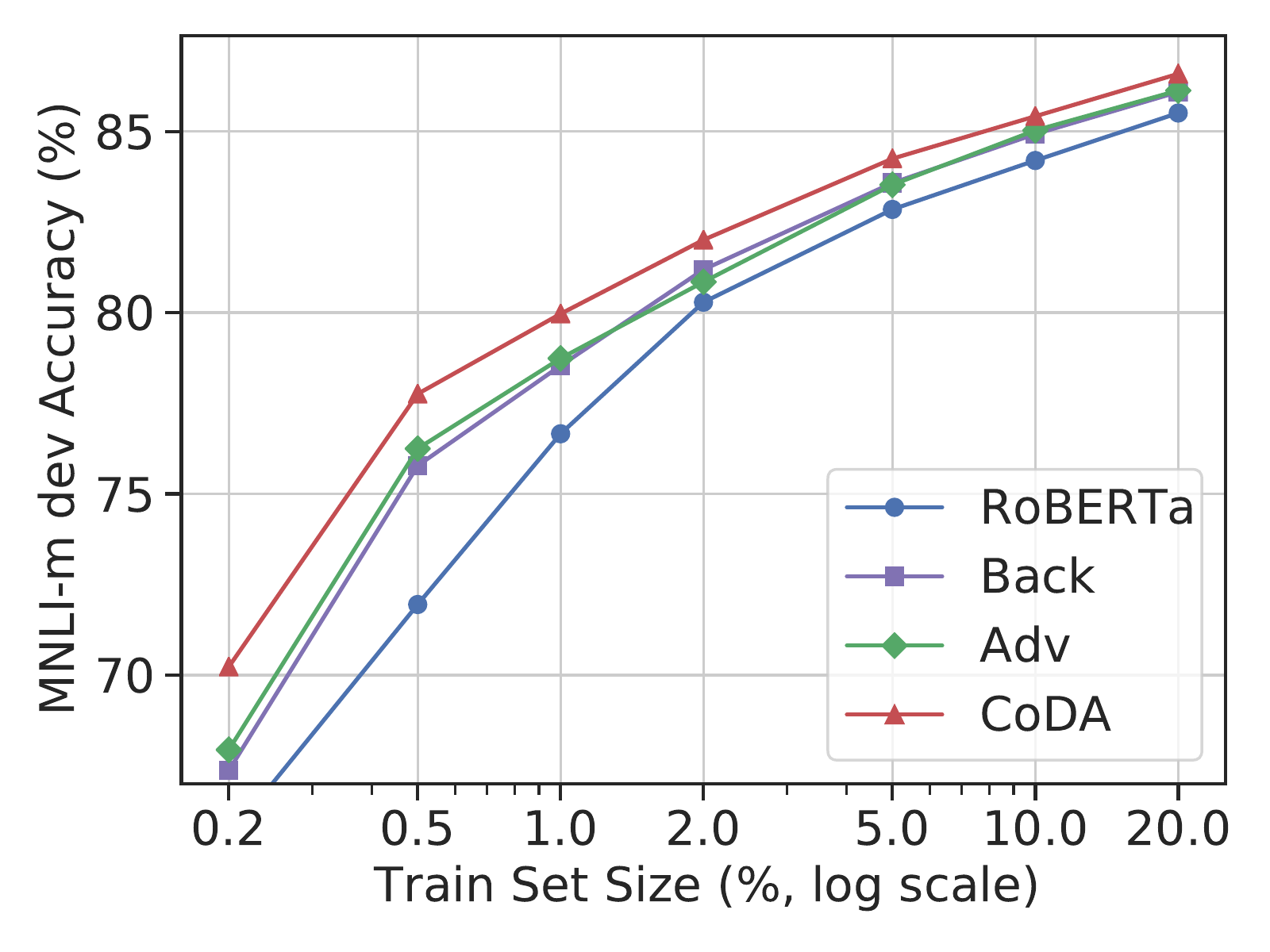}
 \end{subfigure}
  \hspace{15pt}
 \begin{subfigure}[]{0.45\textwidth}
     \centering
     \includegraphics[scale=0.35]{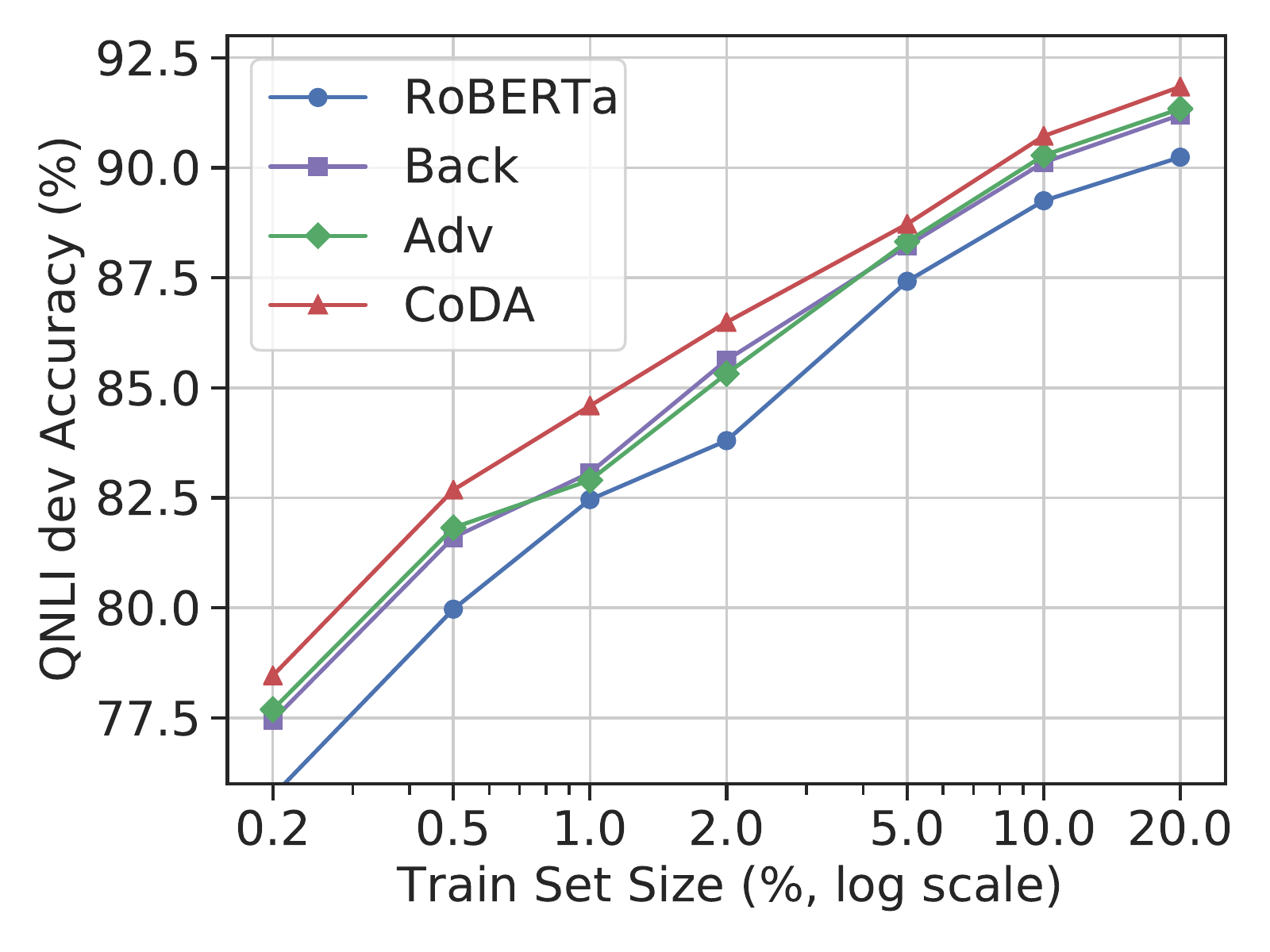}
 \end{subfigure}
 \vspace{-3mm}
    \caption{Low-resource setting experiments on the MNLI (\underline{left}) and QNLI (\underline{right}) dev sets.}
    \vspace{-5mm}
    \label{fig:low}
\end{figure}

\vspace{-3mm}
\subsection{Additional Experiments and Analysis}\label{sec:analysis}
\vspace{-3mm}
\paragraph{Low-resource Setting}

To verify the advantages of CoDA when a smaller number of task-specific data is available, we further conduct a low-resource experiment with the MNLI and QNLI datasets. Concretely, different proportions of training data are sampled and utilized for training. We apply CoDA to RoBERTa-base and compare it with back-translation and adversarial training across various training set sizes. The corresponding results are presented in Fig.~\ref{fig:low}. We observe that back-translation and adversarial training exhibit similar performance across different proportions. More importantly, CoDA demonstrates stronger results consistently, further highlighting its effectiveness with limited training data. 
\vspace{0mm}
\begin{wrapfigure}{r}{0.45\textwidth}
    \centering
    \vspace{0mm}
    \includegraphics[scale=0.32]{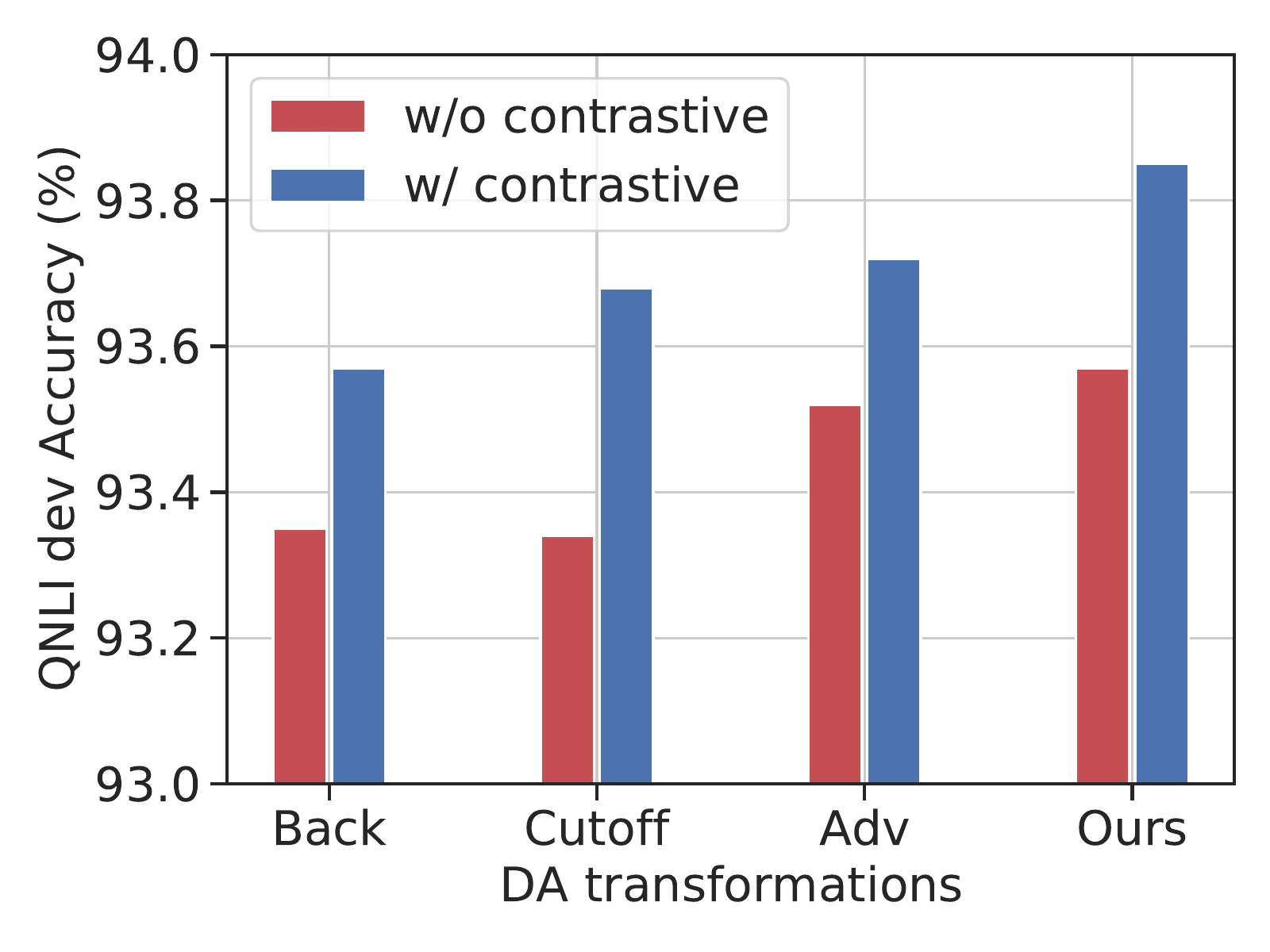}
    \vspace{-4mm}
    \caption{Evaluation of the proposed contrastive objective while applied to different data augmentation approaches.}
    \vspace{-3mm}
    \label{fig:contr}
\end{wrapfigure}
\vspace{-6mm}
\paragraph{The Effectiveness of Contrastive Objective}
To investigate the general applicability of the proposed contrastive regularization objective, we further apply it to different data augmentation methods. The RoBERTa-base model and QNLI dataset are leveraged for this set of experiments, and the results are shown in Fig.~\ref{fig:contr}. We observe that the contrastive learning objective boosts the empirical performance of the resulting algorithm regardless of the data augmentation approaches it is applied to. This further validates our assumption that considering the \emph{global} information among the embeddings of all examples is beneficial for leveraging augmented samples more effectively.


\vspace{-5mm}
\section{Related Work}
\vspace{-3mm}
\paragraph{Data Augmentation in NLP} Different data augmentation approaches have been proposed for text data, such as 
back-translation \citep{sennrich2015improving, edunov2018understanding, xie2019unsupervised}, c-BERT word replacement \citep{wu2019conditional}, mixup \citep{guo2019augmenting, chen-etal-2020-mixtext}, \emph{Cutoff} \citep{shen2020simple}.
Broadly speaking, adversarial training \citep{zhu2019freelb, jiang2019smart} also synthesizes additional examples \emph{via} perturbations at the word embedding layer. 
Although effective, how these data augmentation transformations may be combined together to obtain further improvement has been rarely explored. 
This could be attributed to the fact that a sentence's semantic meanings are quite sensitive to small perturbations.
Consistency-regularized loss \citep{bachman2014learning, rasmus2015semi, laine2016temporal, tarvainen2017mean} is typically employed as the training objective, which \emph{ignores} the global information within the entire dataset.
\vspace{-4mm}
\paragraph{Contrastive Learning} Contrastive methods learn representations by contrasting positive and negative examples, which has demonstrated impressive empirical success in computer vision tasks \citep{henaff2019data, he2020momentum}. Under an unsupervised setting, ontrastive learning approaches learn representation by maximizing mutual information between local-global hidden representations \citep{hjelm2018learning, oord2018representation, henaff2019data}. It can be also leveraged to learn invariant representations by encouraging consensus between augmented samples from the same input \citep{bachman2019learning, tian2019contrastive}. \citet{he2020momentum, wu2018unsupervised} proposes to utilize a memory bank to enable a much larger number of negative samples, which is shown to benefit the transferability of learned representations as well \citep{khosla2020supervised}. Recently, contrastive learning was also employed to improve language model pre-training \citep{iter2020pretraining}.










\vspace{-4mm}
\section{Conclusion}
\vspace{-4mm}
In this paper, we proposed CoDA, a Contrast-enhanced and Diversity promoting data Augmentation framework.
Through extensive experiments, we found that stacking adversarial training over a back-translation module can give rise to more diverse and informative augmented samples. 
Besides, we introduced a specially-designed contrastive loss to incorporate these examples for training in a principled manner. 
Experiments on the GLUE benchmark showed that CoDA consistently improves over several competitive data augmentation and adversarial training baselines. 
Moreover, it is observed that the proposed contrastive objective can be leveraged to improve other data augmentation approaches as well, highlighting the wide applicability of the CoDA framework.

\bibliography{iclr2021_conference}

\begin{thebibliography}{53}
\providecommand{\natexlab}[1]{#1}
\providecommand{\url}[1]{\texttt{#1}}
\expandafter\ifx\csname urlstyle\endcsname\relax
  \providecommand{\doi}[1]{doi: #1}\else
  \providecommand{\doi}{doi: \begingroup \urlstyle{rm}\Url}\fi

\bibitem[Aghajanyan et~al.(2020)Aghajanyan, Shrivastava, Gupta, Goyal,
  Zettlemoyer, and Gupta]{aghajanyan2020better}
Armen Aghajanyan, Akshat Shrivastava, Anchit Gupta, Naman Goyal, Luke
  Zettlemoyer, and Sonal Gupta.
\newblock Better fine-tuning by reducing representational collapse.
\newblock \emph{arXiv preprint arXiv:2008.03156}, 2020.

\bibitem[Bachman et~al.(2014)Bachman, Alsharif, and
  Precup]{bachman2014learning}
Philip Bachman, Ouais Alsharif, and Doina Precup.
\newblock Learning with pseudo-ensembles.
\newblock In \emph{Advances in neural information processing systems}, pp.\
  3365--3373, 2014.

\bibitem[Bachman et~al.(2019)Bachman, Hjelm, and
  Buchwalter]{bachman2019learning}
Philip Bachman, R~Devon Hjelm, and William Buchwalter.
\newblock Learning representations by maximizing mutual information across
  views.
\newblock In \emph{Advances in Neural Information Processing Systems}, pp.\
  15535--15545, 2019.

\bibitem[Berthelot et~al.(2019)Berthelot, Carlini, Goodfellow, Papernot,
  Oliver, and Raffel]{berthelot2019mixmatch}
David Berthelot, Nicholas Carlini, Ian Goodfellow, Nicolas Papernot, Avital
  Oliver, and Colin~A Raffel.
\newblock Mixmatch: A holistic approach to semi-supervised learning.
\newblock In \emph{Advances in Neural Information Processing Systems}, pp.\
  5049--5059, 2019.

\bibitem[Brown et~al.(2020)Brown, Mann, Ryder, Subbiah, Kaplan, Dhariwal,
  Neelakantan, Shyam, Sastry, Askell, et~al.]{brown2020language}
Tom~B Brown, Benjamin Mann, Nick Ryder, Melanie Subbiah, Jared Kaplan, Prafulla
  Dhariwal, Arvind Neelakantan, Pranav Shyam, Girish Sastry, Amanda Askell,
  et~al.
\newblock Language models are few-shot learners.
\newblock \emph{arXiv preprint arXiv:2005.14165}, 2020.

\bibitem[Chen et~al.(2020{\natexlab{a}})Chen, Yang, and
  Yang]{chen-etal-2020-mixtext}
Jiaao Chen, Zichao Yang, and Diyi Yang.
\newblock {M}ix{T}ext: Linguistically-informed interpolation of hidden space
  for semi-supervised text classification.
\newblock In \emph{Proceedings of the 58th Annual Meeting of the Association
  for Computational Linguistics}, pp.\  2147--2157, Online, July
  2020{\natexlab{a}}. Association for Computational Linguistics.
\newblock \doi{10.18653/v1/2020.acl-main.194}.
\newblock URL \url{https://www.aclweb.org/anthology/2020.acl-main.194}.

\bibitem[Chen et~al.(2020{\natexlab{b}})Chen, Kornblith, Norouzi, and
  Hinton]{chen2020simple}
Ting Chen, Simon Kornblith, Mohammad Norouzi, and Geoffrey Hinton.
\newblock A simple framework for contrastive learning of visual
  representations.
\newblock \emph{arXiv preprint arXiv:2002.05709}, 2020{\natexlab{b}}.

\bibitem[Cheng et~al.(2020)Cheng, Jiang, Macherey, and
  Eisenstein]{cheng2020advaug}
Yong Cheng, Lu~Jiang, Wolfgang Macherey, and Jacob Eisenstein.
\newblock Advaug: Robust adversarial augmentation for neural machine
  translation.
\newblock In \emph{Proceedings of the 58th Annual Meeting of the Association
  for Computational Linguistics, {ACL} 2020, Online, July 5-10, 2020}, 2020.

\bibitem[Cubuk et~al.(2018)Cubuk, Zoph, Man{\'e}, Vasudevan, and
  Le]{Cubuk2018AutoAugmentLA}
E.~Cubuk, Barret Zoph, Dandelion Man{\'e}, V.~Vasudevan, and Quoc~V. Le.
\newblock Autoaugment: Learning augmentation policies from data.
\newblock \emph{ArXiv}, abs/1805.09501, 2018.

\bibitem[Cubuk et~al.(2020)Cubuk, Zoph, Shlens, and Le]{cubuk2020randaugment}
Ekin~D Cubuk, Barret Zoph, Jonathon Shlens, and Quoc~V Le.
\newblock Randaugment: Practical automated data augmentation with a reduced
  search space.
\newblock In \emph{Proceedings of the IEEE/CVF Conference on Computer Vision
  and Pattern Recognition Workshops}, pp.\  702--703, 2020.

\bibitem[Devlin et~al.(2019)Devlin, Chang, Lee, and Toutanova]{devlin2018bert}
Jacob Devlin, Ming-Wei Chang, Kenton Lee, and Kristina Toutanova.
\newblock Bert: Pre-training of deep bidirectional transformers for language
  understanding.
\newblock In \emph{Proceedings of the 2019 Conference of the North American
  Chapter of the Association for Computational Linguistics: Human Language
  Technologies, {NAACL-HLT} 2019}, 2019.

\bibitem[DeVries \& Taylor(2017)DeVries and Taylor]{devries2017improved}
Terrance DeVries and Graham~W Taylor.
\newblock Improved regularization of convolutional neural networks with cutout.
\newblock \emph{arXiv preprint arXiv:1708.04552}, 2017.

\bibitem[Edunov et~al.(2018)Edunov, Ott, Auli, and
  Grangier]{edunov2018understanding}
Sergey Edunov, Myle Ott, Michael Auli, and David Grangier.
\newblock Understanding back-translation at scale.
\newblock In \emph{Proceedings of the 2018 Conference on Empirical Methods in
  Natural Language Processing}, 2018.

\bibitem[Gontijo-Lopes et~al.(2020)Gontijo-Lopes, Smullin, Cubuk, and
  Dyer]{gontijo2020affinity}
Raphael Gontijo-Lopes, Sylvia~J Smullin, Ekin~D Cubuk, and Ethan Dyer.
\newblock Affinity and diversity: Quantifying mechanisms of data augmentation.
\newblock \emph{arXiv preprint arXiv:2002.08973}, 2020.

\bibitem[Goodfellow et~al.(2015)Goodfellow, Shlens, and
  Szegedy]{goodfellow2014explaining}
Ian~J Goodfellow, Jonathon Shlens, and Christian Szegedy.
\newblock Explaining and harnessing adversarial examples.
\newblock 2015.

\bibitem[Gretton et~al.(2012)Gretton, Borgwardt, Rasch, Sch{\"o}lkopf, and
  Smola]{gretton2012kernel}
Arthur Gretton, Karsten~M Borgwardt, Malte~J Rasch, Bernhard Sch{\"o}lkopf, and
  Alexander Smola.
\newblock A kernel two-sample test.
\newblock \emph{The Journal of Machine Learning Research}, 13\penalty0
  (1):\penalty0 723--773, 2012.

\bibitem[Guo et~al.(2019)Guo, Mao, and Zhang]{guo2019augmenting}
Hongyu Guo, Yongyi Mao, and Richong Zhang.
\newblock Augmenting data with mixup for sentence classification: An empirical
  study.
\newblock \emph{arXiv preprint arXiv:1905.08941}, 2019.

\bibitem[Hao et~al.(2019)Hao, Dong, Wei, and Xu]{hao2019visualizing}
Yaru Hao, Li~Dong, Furu Wei, and Ke~Xu.
\newblock Visualizing and understanding the effectiveness of bert.
\newblock In \emph{Proceedings of the 2019 Conference on Empirical Methods in
  Natural Language Processing and the 9th International Joint Conference on
  Natural Language Processing, {EMNLP-IJCNLP}}, 2019.

\bibitem[He et~al.(2020)He, Fan, Wu, Xie, and Girshick]{he2020momentum}
Kaiming He, Haoqi Fan, Yuxin Wu, Saining Xie, and Ross Girshick.
\newblock Momentum contrast for unsupervised visual representation learning.
\newblock In \emph{Proceedings of the IEEE/CVF Conference on Computer Vision
  and Pattern Recognition}, pp.\  9729--9738, 2020.

\bibitem[H{\'e}naff et~al.(2019)H{\'e}naff, Srinivas, De~Fauw, Razavi, Doersch,
  Eslami, and Oord]{henaff2019data}
Olivier~J H{\'e}naff, Aravind Srinivas, Jeffrey De~Fauw, Ali Razavi, Carl
  Doersch, SM~Eslami, and Aaron van~den Oord.
\newblock Data-efficient image recognition with contrastive predictive coding.
\newblock \emph{arXiv preprint arXiv:1905.09272}, 2019.

\bibitem[Hendrycks et~al.(2020)Hendrycks, Mu, Cubuk, Zoph, Gilmer, and
  Lakshminarayanan]{hendrycks2019augmix}
Dan Hendrycks, Norman Mu, Ekin~D Cubuk, Barret Zoph, Justin Gilmer, and Balaji
  Lakshminarayanan.
\newblock Augmix: A simple data processing method to improve robustness and
  uncertainty.
\newblock In \emph{8th International Conference on Learning Representations,
  {ICLR}}, 2020.

\bibitem[Hjelm et~al.(2019)Hjelm, Fedorov, Lavoie-Marchildon, Grewal, Bachman,
  Trischler, and Bengio]{hjelm2018learning}
R~Devon Hjelm, Alex Fedorov, Samuel Lavoie-Marchildon, Karan Grewal, Phil
  Bachman, Adam Trischler, and Yoshua Bengio.
\newblock Learning deep representations by mutual information estimation and
  maximization.
\newblock In \emph{7th International Conference on Learning Representations,
  {ICLR}}, 2019.

\bibitem[Hoang et~al.(2018)Hoang, Koehn, Haffari, and Cohn]{hoang2018iterative}
Vu~Cong~Duy Hoang, Philipp Koehn, Gholamreza Haffari, and Trevor Cohn.
\newblock Iterative back-translation for neural machine translation.
\newblock In \emph{Proceedings of the 2nd Workshop on Neural Machine
  Translation and Generation}, pp.\  18--24, 2018.

\bibitem[Hu et~al.(2019)Hu, Tan, Salakhutdinov, Mitchell, and
  Xing]{hu2019learning}
Zhiting Hu, Bowen Tan, Russ~R Salakhutdinov, Tom~M Mitchell, and Eric~P Xing.
\newblock Learning data manipulation for augmentation and weighting.
\newblock In \emph{Advances in Neural Information Processing Systems}, pp.\
  15764--15775, 2019.

\bibitem[Iter et~al.(2020)Iter, Guu, Lansing, and
  Jurafsky]{iter2020pretraining}
Dan Iter, Kelvin Guu, Larry Lansing, and Dan Jurafsky.
\newblock Pretraining with contrastive sentence objectives improves discourse
  performance of language models.
\newblock In \emph{Proceedings of the 58th Annual Meeting of the Association
  for Computational Linguistics, {ACL}}, 2020.

\bibitem[Jiang et~al.(2020)Jiang, He, Chen, Liu, Gao, and Zhao]{jiang2019smart}
Haoming Jiang, Pengcheng He, Weizhu Chen, Xiaodong Liu, Jianfeng Gao, and Tuo
  Zhao.
\newblock Smart: Robust and efficient fine-tuning for pre-trained natural
  language models through principled regularized optimization.
\newblock In \emph{Proceedings of the 58th Annual Meeting of the Association
  for Computational Linguistics, {ACL}}, 2020.

\bibitem[Kannan et~al.(2018)Kannan, Kurakin, and
  Goodfellow]{kannan2018adversarial}
Harini Kannan, Alexey Kurakin, and Ian Goodfellow.
\newblock Adversarial logit pairing.
\newblock \emph{arXiv preprint arXiv:1803.06373}, 2018.

\bibitem[Khosla et~al.(2020)Khosla, Teterwak, Wang, Sarna, Tian, Isola,
  Maschinot, Liu, and Krishnan]{khosla2020supervised}
Prannay Khosla, Piotr Teterwak, Chen Wang, Aaron Sarna, Yonglong Tian, Phillip
  Isola, Aaron Maschinot, Ce~Liu, and Dilip Krishnan.
\newblock Supervised contrastive learning.
\newblock \emph{arXiv preprint arXiv:2004.11362}, 2020.

\bibitem[Kingma \& Ba(2014)Kingma and Ba]{kingma2014adam}
Diederik~P Kingma and Jimmy Ba.
\newblock Adam: A method for stochastic optimization.
\newblock In \emph{3rd International Conference on Learning Representations,
  {ICLR} 2015}, 2014.

\bibitem[Kumar et~al.(2019)Kumar, Bhattamishra, Bhandari, and
  Talukdar]{kumar2019submodular}
Ashutosh Kumar, Satwik Bhattamishra, Manik Bhandari, and Partha Talukdar.
\newblock Submodular optimization-based diverse paraphrasing and its
  effectiveness in data augmentation.
\newblock In \emph{Proceedings of the 2019 Conference of the North American
  Chapter of the Association for Computational Linguistics: Human Language
  Technologies}, 2019.

\bibitem[Laine \& Aila(2017)Laine and Aila]{laine2016temporal}
Samuli Laine and Timo Aila.
\newblock Temporal ensembling for semi-supervised learning.
\newblock In \emph{5th International Conference on Learning Representations,
  {ICLR} 2017, Toulon, France, April 24-26, 2017, Conference Track
  Proceedings}. OpenReview.net, 2017.

\bibitem[Liu et~al.(2020)Liu, Cheng, He, Chen, Wang, Poon, and
  Gao]{liu2020adversarial}
Xiaodong Liu, Hao Cheng, Pengcheng He, Weizhu Chen, Yu~Wang, Hoifung Poon, and
  Jianfeng Gao.
\newblock Adversarial training for large neural language models.
\newblock \emph{arXiv preprint arXiv:2004.08994}, 2020.

\bibitem[Liu et~al.(2019)Liu, Ott, Goyal, Du, Joshi, Chen, Levy, Lewis,
  Zettlemoyer, and Stoyanov]{liu2019roberta}
Yinhan Liu, Myle Ott, Naman Goyal, Jingfei Du, Mandar Joshi, Danqi Chen, Omer
  Levy, Mike Lewis, Luke Zettlemoyer, and Veselin Stoyanov.
\newblock Roberta: A robustly optimized bert pretraining approach.
\newblock \emph{arXiv preprint arXiv:1907.11692}, 2019.

\bibitem[Miyato et~al.(2018)Miyato, Maeda, Koyama, and
  Ishii]{miyato2018virtual}
Takeru Miyato, Shin-ichi Maeda, Masanori Koyama, and Shin Ishii.
\newblock Virtual adversarial training: a regularization method for supervised
  and semi-supervised learning.
\newblock \emph{IEEE transactions on pattern analysis and machine
  intelligence}, 41\penalty0 (8):\penalty0 1979--1993, 2018.

\bibitem[Oord et~al.(2018)Oord, Li, and Vinyals]{oord2018representation}
Aaron van~den Oord, Yazhe Li, and Oriol Vinyals.
\newblock Representation learning with contrastive predictive coding.
\newblock \emph{arXiv preprint arXiv:1807.03748}, 2018.

\bibitem[Raffel et~al.(2019)Raffel, Shazeer, Roberts, Lee, Narang, Matena,
  Zhou, Li, and Liu]{raffel2019exploring}
Colin Raffel, Noam Shazeer, Adam Roberts, Katherine Lee, Sharan Narang, Michael
  Matena, Yanqi Zhou, Wei Li, and Peter~J Liu.
\newblock Exploring the limits of transfer learning with a unified text-to-text
  transformer.
\newblock \emph{arXiv preprint arXiv:1910.10683}, 2019.

\bibitem[Rasmus et~al.(2015)Rasmus, Berglund, Honkala, Valpola, and
  Raiko]{rasmus2015semi}
Antti Rasmus, Mathias Berglund, Mikko Honkala, Harri Valpola, and Tapani Raiko.
\newblock Semi-supervised learning with ladder networks.
\newblock In \emph{Advances in neural information processing systems}, pp.\
  3546--3554, 2015.

\bibitem[Sennrich et~al.(2016)Sennrich, Haddow, and
  Birch]{sennrich2015improving}
Rico Sennrich, Barry Haddow, and Alexandra Birch.
\newblock Improving neural machine translation models with monolingual data.
\newblock In \emph{Proceedings of the 54th Annual Meeting of the Association
  for Computational Linguistics, {ACL}}, 2016.

\bibitem[Shen et~al.(2020)Shen, Zheng, Shen, Qu, and Chen]{shen2020simple}
Dinghan Shen, Mingzhi Zheng, Yelong Shen, Yanru Qu, and Weizhu Chen.
\newblock A simple but tough-to-beat data augmentation approach for natural
  language understanding and generation, 2020.

\bibitem[Shen et~al.(2018)Shen, Qu, Zhang, and Yu]{shen2017wasserstein}
Jian Shen, Yanru Qu, Weinan Zhang, and Yong Yu.
\newblock Wasserstein distance guided representation learning for domain
  adaptation.
\newblock In \emph{Proceedings of the Thirty-Second {AAAI} Conference on
  Artificial Intelligence, (AAAI-18)}, 2018.

\bibitem[Sohn et~al.(2020)Sohn, Berthelot, Li, Zhang, Carlini, Cubuk, Kurakin,
  Zhang, and Raffel]{sohn2020fixmatch}
Kihyuk Sohn, David Berthelot, Chun-Liang Li, Zizhao Zhang, Nicholas Carlini,
  Ekin~D Cubuk, Alex Kurakin, Han Zhang, and Colin Raffel.
\newblock Fixmatch: Simplifying semi-supervised learning with consistency and
  confidence.
\newblock \emph{arXiv preprint arXiv:2001.07685}, 2020.

\bibitem[Sun et~al.(2019)Sun, Qiu, Xu, and Huang]{sun2019fine}
Chi Sun, Xipeng Qiu, Yige Xu, and Xuanjing Huang.
\newblock How to fine-tune bert for text classification?
\newblock In \emph{China National Conference on Chinese Computational
  Linguistics}, pp.\  194--206. Springer, 2019.

\bibitem[Tarvainen \& Valpola(2017)Tarvainen and Valpola]{tarvainen2017mean}
Antti Tarvainen and Harri Valpola.
\newblock Mean teachers are better role models: Weight-averaged consistency
  targets improve semi-supervised deep learning results.
\newblock In \emph{Advances in neural information processing systems}, pp.\
  1195--1204, 2017.

\bibitem[Tian et~al.(2019)Tian, Krishnan, and Isola]{tian2019contrastive}
Yonglong Tian, Dilip Krishnan, and Phillip Isola.
\newblock Contrastive multiview coding.
\newblock \emph{arXiv preprint arXiv:1906.05849}, 2019.

\bibitem[Wang et~al.(2018)Wang, Singh, Michael, Hill, Levy, and
  Bowman]{wang2018glue}
Alex Wang, Amanpreet Singh, Julian Michael, Felix Hill, Omer Levy, and Samuel~R
  Bowman.
\newblock Glue: A multi-task benchmark and analysis platform for natural
  language understanding.
\newblock \emph{EMNLP 2018}, pp.\  353, 2018.

\bibitem[Wei \& Zou(2019)Wei and Zou]{wei2019eda}
Jason Wei and Kai Zou.
\newblock Eda: Easy data augmentation techniques for boosting performance on
  text classification tasks.
\newblock In \emph{Proceedings of the 2019 Conference on Empirical Methods in
  Natural Language Processing and the 9th International Joint Conference on
  Natural Language Processing, {EMNLP-IJCNLP}}, 2019.

\bibitem[Wu et~al.(2019)Wu, Lv, Zang, Han, and Hu]{wu2019conditional}
Xing Wu, Shangwen Lv, Liangjun Zang, Jizhong Han, and Songlin Hu.
\newblock Conditional bert contextual augmentation.
\newblock In \emph{International Conference on Computational Science}, pp.\
  84--95. Springer, 2019.

\bibitem[Wu et~al.(2018)Wu, Xiong, Yu, and Lin]{wu2018unsupervised}
Zhirong Wu, Yuanjun Xiong, Stella~X Yu, and Dahua Lin.
\newblock Unsupervised feature learning via non-parametric instance
  discrimination.
\newblock In \emph{Proceedings of the IEEE Conference on Computer Vision and
  Pattern Recognition}, pp.\  3733--3742, 2018.

\bibitem[Xie et~al.(2019)Xie, Dai, Hovy, Luong, and Le]{xie2019unsupervised}
Qizhe Xie, Zihang Dai, Eduard Hovy, Minh-Thang Luong, and Quoc~V Le.
\newblock Unsupervised data augmentation for consistency training.
\newblock \emph{arXiv preprint arXiv:1904.12848}, 2019.

\bibitem[Yu et~al.(2018)Yu, Dohan, Luong, Zhao, Chen, Norouzi, and
  Le]{yu2018qanet}
Adams~Wei Yu, David Dohan, Minh-Thang Luong, Rui Zhao, Kai Chen, Mohammad
  Norouzi, and Quoc~V Le.
\newblock Qanet: Combining local convolution with global self-attention for
  reading comprehension.
\newblock In \emph{6th International Conference on Learning Representations,
  {ICLR} 2018}, 2018.

\bibitem[Zhang et~al.(2017)Zhang, Cisse, Dauphin, and
  Lopez-Paz]{zhang2017mixup}
Hongyi Zhang, Moustapha Cisse, Yann~N Dauphin, and David Lopez-Paz.
\newblock mixup: Beyond empirical risk minimization.
\newblock \emph{arXiv preprint arXiv:1710.09412}, 2017.

\bibitem[Zheng et~al.(2016)Zheng, Song, Leung, and
  Goodfellow]{zheng2016improving}
Stephan Zheng, Yang Song, Thomas Leung, and Ian Goodfellow.
\newblock Improving the robustness of deep neural networks via stability
  training.
\newblock In \emph{Proceedings of the ieee conference on computer vision and
  pattern recognition}, pp.\  4480--4488, 2016.

\bibitem[Zhu et~al.(2020)Zhu, Cheng, Gan, Sun, Goldstein, and
  Liu]{zhu2019freelb}
Chen Zhu, Yu~Cheng, Zhe Gan, Siqi Sun, Thomas Goldstein, and Jingjing Liu.
\newblock Freelb: Enhanced adversarial training for language understanding.
\newblock In \emph{8th International Conference on Learning Representations,
  {ICLR}}, 2020.

\end{thebibliography}
\bibliographystyle{iclr2021_conference}

\appendix

\section{Data Augmentation Details}\label{ap:da}

We select the following representative data augmentation operations as basic building blocks of our data augmentation module. We denote $\vx_i = [x_{i,1}, \dots, x_{i,l}]$ as the input text sequence, and $\ve_i = [\ve_{i,1}, \dots, \ve_{i,l}]$ as corresponding embedding vectors.
\begin{itemize}[leftmargin=*]
\item Back translation is widely applied in machine translation (MT) \citep{sennrich2015improving, hoang2018iterative, edunov2018understanding}, and is introduced to text classification recently \citep{xie2019unsupervised}. Back-Trans uses 2 MT models to translate the input example to another pivot language, and then translate it back, $\vx_i \rightarrow \text{Pivot Language} \rightarrow \vx'_i$.
    
\item C-BERT Word Replacement \citep{wu2019conditional} is a representative of the word replacement augmentation family. C-BERT pretrains a conditional BERT model to learn contextualized representation  $P(\rx_{j} | [x_{i,1} \dots x_{i,j-1} \text{[MASK]} x_{i,j+1} \dots x_{i,l}], y_i)$ conditioning on classes. This method then randomly substitutes words of $\vx$ to obtain $\vx'$ ($[x_{i,1} \dots x'_{i,j} \dots x_{i,l}]$)\footnote{EDA \citep{wei2019eda} uses synonym replacement, another word replacement technique. We choose C-BERT for this family to take the advantages of contextual representation.}.
    
\item Cutoff \citep{devries2017improved} randomly drops units in a continuous span on the input, while \citet{shen2020simple} adapts this method to text embeddings. For input embeddings $\ve_i$, this method randomly set a continuous span of elements to 0s, $\ve'_i = [\ve_{i,1} \dots \ve_{i,j-1}, 0 \dots 0, \ve_{i,j+w} \dots \ve_{i,l}]$, where the window size $w \propto l$, and the start position $j \in [1, l-w]$ is randomly selected. For transformer encoders that involve position embeddings, we also set input mask as 0s at corresponding positions.
    
\item Mixup \citep{zhang2017mixup} interpolates two image as well as their labels. \citet{guo2019augmenting} borrows this method to text. For 2 input embeddings $(\ve_i, \ve_j)$, mixup interpolates the embedding vectors $\ve'_i = a \ve_i + (1-a) \ve_j$ where $a$ is sampled from a Beta distribution. Also, the labels are interpolated for the augmented sample $y'_i = a y_i + (1-a) y_j$.
    
\item Adversarial training generates adversarial examples for input embeddings, simply, $\ve'_i = \argmax_{\Vert \ve_i - \ve'_i \Vert \le 1} \mathcal{L}(f(\ve'_i), y_i)$. We mainly follow the implementation of \citet{zhu2019freelb}. Besides, when computing the adversarial example $\ve'_i$, the dropout variables are recorded and reused later when encoding $\ve'_i$.
\end{itemize}

Maximum mean discrepancy (MMD) \citep{gretton2012kernel} is a widely used discrepancy measure for 2 distributions. We adopt the multi-kernel MMD implementation based on \citet{shen2017wasserstein}\footnote{https://github.com/RockySJ/WDGRL}, to quantify the distance of data distributions before and after DA transformations.

\section{Dataset Details}\label{ap:data}

The datasets and statistics are summarized in Table~\ref{tab:data}.
\begin{table}[h]
    \centering
    \begin{tabular}{l|c|c|c|c|c|c|c} \hline
        \textbf{Corpus} & Task & \tabincell{c}{Sentence \\ Pair} & \#Train & \#Dev & \#Test & \#Class & Metrics \\ \hline \hline
        MNLI & NLI & \checkmark & 393k & 20k & 20k & 3 & Accuracy \\ \hline
        QQP & Paraphrase & \checkmark & 364k & 40k & 391k & 2 & Accuracy/F1 \\ \hline 
        QNLI & QA/NLI & \checkmark & 108k & 5.7k & 5.7k & 2 & Accuracy \\ \hline
        SST & Sentiment & $\times$ & 67k & 872 & 1.8k & 2 & Accuracy \\ \hline
        MRPC & Paraphrase & \checkmark & 3.7k & 408 & 1.7k & 2 & Accuracy/F1 \\ \hline
        CoLA & Acceptability & $\times$ & 8.5k & 1k & 1k & 2 & Matthews corr \\ \hline
        RTE & NLI & \checkmark & 2.5k & 276 & 3k & 2 & Accuracy \\ \hline
        STS-B & Similarity & \checkmark & 7k & 1.5k & 1.4k & - & \tabincell{c}{Pearson/Spe-\\arman corr} \\ \hline
    \end{tabular}
    \caption{GLUE benchmark summary.}
    \label{tab:data}
\end{table}

\section{Implementation Details}\label{ap:implement}

Our implementation is based on RoBERTa \citep{liu2019roberta}. We use ADAM \citep{kingma2014adam} as our optimizer. We follow the hyper-parameter study of RoBERTa and set as default the following parameters: batch size (32), learning rate (1e-5), epochs (5), warmup ratio (0.06), weight decay (0.1) and we keep other parameters unchanged with RoBERTa. For Back-Trans, we use the en-de single models trained on WMT19 and released in FairSeq. More specifically, we use beam search (beam size = 5) and keep only the top-1 hypothesis. We slightly tune Adversarial parameters on MNLI based on FreeLB and fix them on other datasets, since adversarial training is not our focus. For contrastive regularization, we implement based on MoCo. 
In GLUE evaluation, we mainly tune the weights of 3 regularization terms, $\alpha \in [0, 1], \beta \in [0, 3], \lambda \in [0, 0.03]$ (Eq.~\ref{eq:stack_loss}, \ref{eq:objective}). Besides, for smaller tasks (MRPC, CoLA, RTE, STS-B), we use the best performed MNLI model to initialize their parameters\footnote{RoBERTa: https://github.com/huggingface/transformers. FairSeq: https://github.com/pytorch/fairseq. FreeLB: https://github.com/zhuchen03/FreeLB. MoCo: https://github.com/facebookresearch/moco. We will release our model and code for further study.}. 


\end{document}